\documentclass[10pt,twocolumn,letterpaper]{article}
\usepackage[T1]{fontenc}
\usepackage[pagenumbers]{cvpr}

\setlist[itemize]{noitemsep,leftmargin=*,topsep=2pt,partopsep=0pt,parsep=0pt}

\makeatletter
\patchcmd{\@maketitle}{\vspace*{24pt}}{\vspace*{12pt}}{}{}
\patchcmd{\@maketitle}{\begin{tabular}[t]{c}}{\begin{tabular}[t]{@{}c@{}}}{}{}
\makeatother

\definecolor{cvprblue}{rgb}{0.21,0.49,0.74}
\usepackage[breaklinks,colorlinks,allcolors=cvprblue]{hyperref}
\hypersetup{
    pdftitle={PestWeave: Linking Semantic Modeling and Evidence-Guided Text Modulation through Visual Prototypes for Fine-Grained Pest Learning},
    pdfauthor={Xueheng Li, Yu Wang, Yuxiang Shen, Wen Xue, Rui Li, Jie Zhang, Chengjun Xie},
    pdfsubject={Fine-grained multimodal pest recognition},
    pdfkeywords={pest recognition, vision-language learning, visual prototypes, RWKV}
}

\title{PestWeave: Linking Semantic Modeling and Evidence-Guided Text Modulation through Visual Prototypes for Fine-Grained Pest Learning}
\author{%
    Xueheng Li$^{1,2}$,\enspace
    Yu Wang$^{1,2}$,\enspace
    Yuxiang Shen$^{1,2}$,\enspace
    Wen Xue$^{4}$,\\
    Rui Li$^{1}$,\enspace
    Jie Zhang$^{1,3}$,\enspace
    Chengjun Xie$^{1,3\dagger}$\\[0.1em]
    {\fontsize{12}{13.2}\selectfont
    \renewcommand{\arraystretch}{0.95}%
    \begin{tabular}[t]{@{}c@{}}
        $^{1}$Institute of Intelligent Machines, Hefei Institutes of Physical Science, Chinese Academy of Sciences\\
        $^{2}$University of Science and Technology of China\\
        $^{3}$Zhongke Hefei Institute of Technology Innovation Engineering\\
        $^{4}$South China University of Technology
    \end{tabular}}\\[0.1em]
    {\ttfamily\small lixueheng@mail.ustc.edu.cn, \{lirui,zhangjie,cjxie\}@iim.ac.cn}%
}

\newcommand{\arxivAuthorFootnotes}{%
    \begingroup
    \renewcommand{\thefootnote}{\fnsymbol{footnote}}%
    \footnotetext[2]{Corresponding author.}%
    \endgroup
}

\newcommand{\pestAblationTableWidth}{\columnwidth}

\begin{document}
\maketitle
\arxivAuthorFootnotes
\begin{abstract}
Agricultural pests threaten crop yield and quality, making accurate pest recognition essential for targeted management and sustainable agriculture. Their diverse appearances complicate recognition and motivate multimodal approaches incorporating complementary species knowledge from text. However, complex backgrounds can obscure discriminative visual cues, while textual descriptions may contain information unsupported by the current image, limiting effective cross-modal fusion. To address these challenges, we propose PestWeave, a multimodal framework connecting content-adaptive visual organization with text fusion through visual prototypes. Its visual pathway augments a Receptance Weighted Key Value (RWKV) architecture with semantic token grouping and three types of auxiliary tokens, enabling content-adaptive scanning and contextual modeling while preserving spatial structure. Feature-similar tokens are reordered into adjacent scanning positions, while cluster prototype, global, and register tokens provide group-level summaries, global context, and additional interaction capacity, respectively. These prototypes also guide text modulation by assessing visual support for semantic components and regulating their relative contributions and overall fusion strength, reducing interference from unsupported descriptions. Extensive experiments on multiple pest recognition datasets demonstrate the effectiveness of PestWeave and its potential for practical pest management.
\end{abstract}
\section{Introduction}
\label{sec:intro}
\begin{figure}[ht]
    \centering
    \includegraphics[width=\columnwidth]{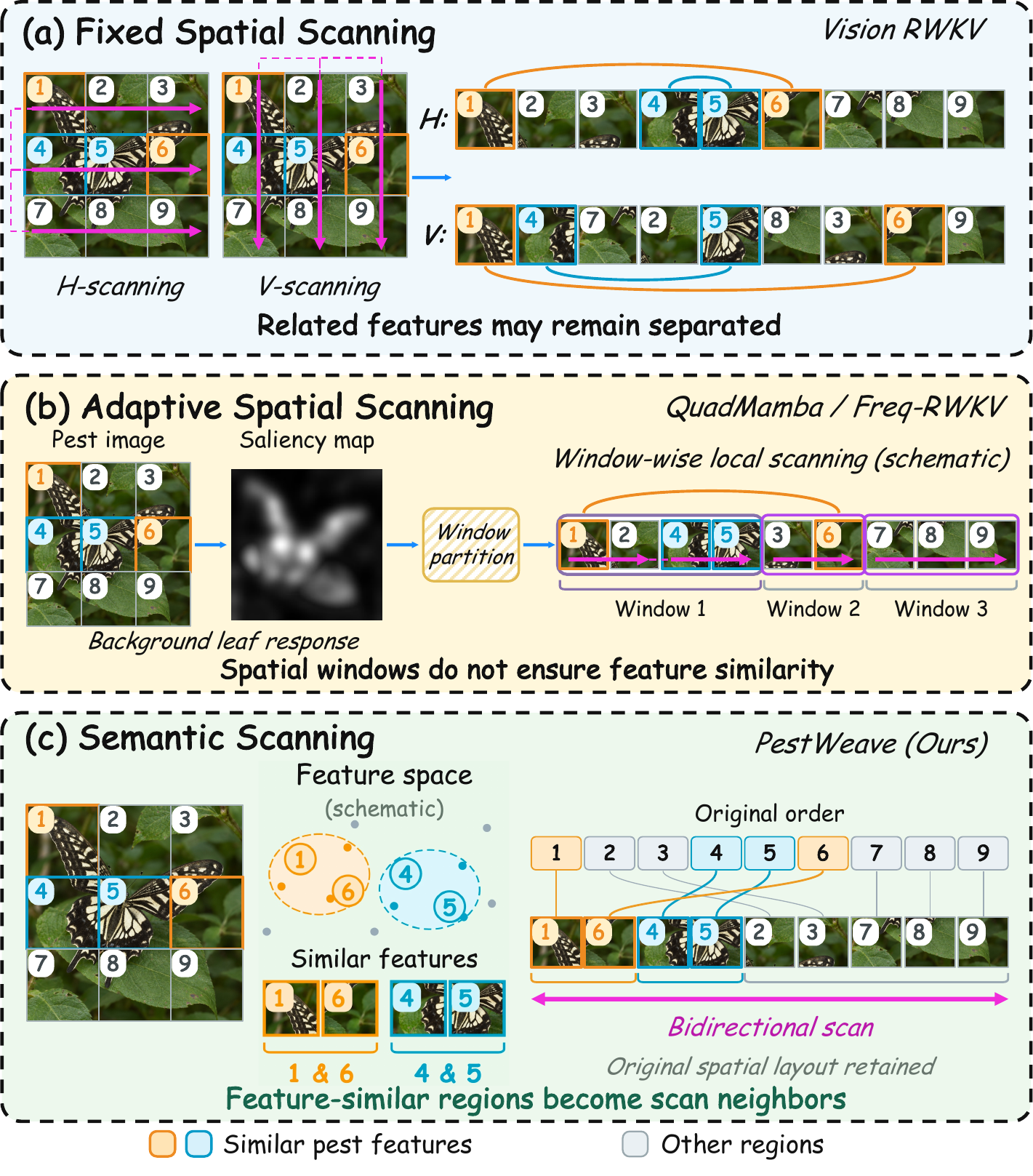}
    \caption{ (a) Fixed spatial scanning may separate visually similar regions. (b) Saliency-guided adaptive scanning is susceptible to interference from salient background regions. (c) Our semantic scanning brings feature-similar regions into neighboring sequence positions while preserving the original spatial layout.
    }
    \label{fig:motivation}
\end{figure}

Accurate crop pest recognition is essential for targeted pest management and sustainable agriculture \cite{savary2019global, ma2025crop, amiri2026artificial}. However, substantial appearance variations within species across developmental stages and viewpoints, subtle morphological differences between species, and complex field backgrounds make fine-grained pest recognition particularly challenging. To complement visual representations, recent multimodal approaches incorporate species-specific textual knowledge to facilitate the discrimination of visually similar pests \cite{li2026pest, li2026pestvl, truong2025insect}. Despite these advances, effective multimodal recognition faces two closely related challenges: extracting discriminative visual features from complex scenes and determining whether the semantic knowledge conveyed by text is supported by the current image. Addressing both challenges is thus critical for reliable fine-grained pest recognition, where species discrimination depends on capturing subtle morphological characteristics while reducing interference from textual semantics unsupported by the current image.

Visual modeling requires capturing fine-grained morphological details while maintaining global contextual interactions and computational efficiency. Receptance Weighted Key Value (RWKV) \cite{peng2023rwkv, duan2025visionrwkv} provides an efficient mechanism for modeling long-range dependencies with linear complexity. However, conventional spatial scanning follows fixed horizontal or vertical orders without considering content similarity, leaving visually related regions potentially distant in the scanning sequence \cite{guo2025mambairv2}. As illustrated in Fig.~\ref{fig:motivation}, existing saliency-guided strategies \cite{li2026pestvl, li2025freq} improve local feature granularity but rely on low-level responses that may also highlight texture-rich backgrounds rather than discriminative pest structures. These limitations motivate content-adaptive organization that strengthens interactions among visually related regions. Meanwhile, local cues such as markings and wing venation must be interpreted in the context of overall morphology, making group-level aggregation alone insufficient. Effective visual modeling should therefore integrate semantic relationships with spatial structure and image-wide context while reducing interference from irrelevant responses.

Textual descriptions provide complementary knowledge about pest morphology and species characteristics, yet their semantic content is not equally applicable to every image. Variations in viewpoint, partial occlusion, and image quality may render certain described attributes unobservable, while generic or background-related information can introduce irrelevant semantics \cite{ma2025multi, lee2025enhancing}. Although vision--language interaction enables adaptive feature fusion \cite{li2026pestvl, li2026clarif}, conventional cross-attention primarily models the relative importance of textual components and does not explicitly account for cases where the available visual evidence provides limited support for the description as a whole. Consequently, effective fusion requires not only identifying applicable semantic components but also regulating how strongly textual information modifies visual features at different spatial positions. This motivates an evidence-aware mechanism that can distinguish visually supported semantics from unsupported descriptions and adaptively control their contributions to recognition.

To address these challenges, we propose PestWeave, a multimodal framework that connects content-adaptive visual modeling with evidence-guided text modulation through shared visual prototypes. In the visual pathway, we introduce an Locality-Sensitive Hashing (LSH)-initialized \cite{datar2004locality} token grouping mechanism with spherical clustering refinement, which organizes feature-similar tokens into adjacent positions for semantic scanning. This content-adaptive pathway complements conventional spatial scanning, with semantic features restored to their original spatial positions before residual fusion to preserve spatial structure. We further incorporate cluster prototype, global, and learnable register tokens into the semantic scan to enrich group-level interactions and global contextual modeling. Importantly, the resulting visual prototypes also provide compact representations of image content for assessing the applicability of textual semantics. In the text pathway, a frozen CLIP encoder and query-based compression produce latent semantic tokens from the paired caption. We estimate their visual support by matching them against cluster prototypes and a learnable null-evidence key, allowing semantic components without corresponding visual evidence to receive reduced contributions. The resulting evidence estimates guide both text attention reweighting and spatially gated injection, jointly regulating semantic contributions and the magnitude of text updates at individual visual positions. By using visual prototypes for both semantic feature organization and cross-modal modulation, PestWeave establishes a direct connection between discriminative visual modeling and evidence-aware text fusion.

In summary, our contributions are as follows:

\begin{itemize}
    \item We propose PestWeave, a visual evidence-guided framework for fine-grained multimodal pest learning that connects content-adaptive visual modeling with text fusion through cluster-level visual prototypes.
    \item We develop an LSH-assisted token clustering and semantic scanning mechanism incorporating cluster prototype, global, and register tokens. It enhances interactions among related regions and contextual modeling while preserving spatial structure and alleviating interference from complex backgrounds.
    \item We introduce a text modulation mechanism supported by visual prototypes that combines prototype matching with null-evidence modeling to jointly regulate the relative contributions of semantic components and the overall strength of text injection, reducing interference from semantics lacking visual support.
    \item Extensive experiments on multiple datasets validate the effectiveness of PestWeave. Ablation studies and visualizations further demonstrate the contributions of individual components and the complementarity between visual organization and text modulation.
\end{itemize}

\section{Related Work}
\label{sec:related}

\subsection{Visual and Multimodal Pest Learning}
Deep learning has been widely adopted for fine-grained agricultural pest recognition to address subtle morphological similarities and complex field backgrounds.
Early work by Li et al.~\cite{li2020croppest} utilized convolutional networks for pest recognition in natural scenes.
Subsequent works optimize feature learning against appearance variations and environmental noise.
DFL~\cite{hu2024dfl}, InsectMamba~\cite{wang2025insectmamba}, and HIVE~\cite{kim2026hive} leverage causal constraints, state-space modeling, and hierarchical pre-training respectively.
These methods focus on learning discriminative visual representations.
However, most of them rely solely on visual information and fail to exploit descriptive biological semantic knowledge for ambiguous samples. Multimodal approaches introduce textual information as auxiliary cues.
Insect-Foundation~\cite{nguyen2024insectfoundation} constructs the Insect-1M dataset with description-consistency supervision.
AgroBench~\cite{shinoda2025agrobench} evaluates vision-language models on pest tasks and highlights persistent challenges in fine-grained recognition.
PestVL-Net~\cite{li2026pestvl} and Pest-MoE~\cite{li2026pestmoe} further enable vision-text interaction and modal gating fusion.
Although existing works augment visual features with text, they lack explicit modeling of the alignment between textual descriptions and visible image content.
In this work, we dynamically learn text weights from cluster-level visual evidence, coupling discriminative visual features with fine-grained semantic weighting.

\subsection{Recurrent Weighted Key Value}
RWKV~\cite{peng2023rwkv} supports linear-complexity sequence aggregation, parallel training, and recurrent inference, offering a promising paradigm for efficient long-range visual modeling.
Vision-RWKV~\cite{duan2025visionrwkv} extends RWKV to visual tasks, and follow-up works such as ChangeRWKV~\cite{yang2026changerwkv} and CLUIE~\cite{jiang2026cluie} validate its potential on remote sensing and underwater image tasks.
Nevertheless, most of these models adopt fixed spatial scanning orders and struggle to group discriminative tokens with similar appearance but scattered spatial locations.
For pest recognition, PestVL-Net~\cite{li2026pestvl} adopts saliency-guided windows to allocate computation.
However, low-level saliency signals are prone to interference from textured backgrounds and lack pest-specific semantic grouping.
To address this limitation, we propose a content-driven token scanning strategy with LSH initialization and spherical clustering refinement.
The learned cluster prototypes support both RWKV visual modeling and cross-modal text fusion.

\section{Method}
\label{sec:method}
\begin{figure*}[ht]
    \centering
    \includegraphics[width=0.97\textwidth]{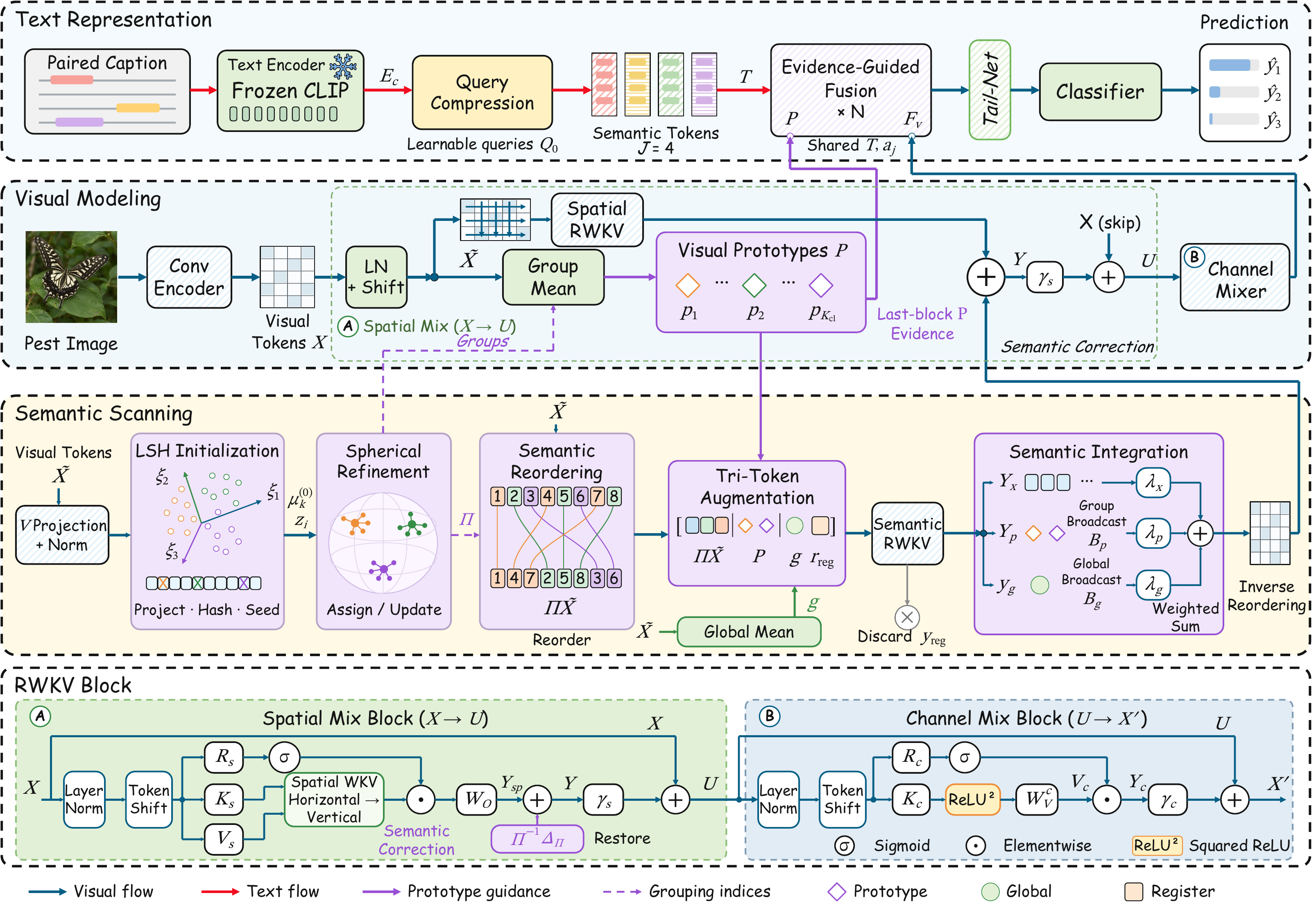}
    \caption{{Overview of the proposed PestWeave framework.}
    The visual branch integrates spatial modeling with
    semantic scanning and tri-token augmentation.
    Visual prototypes further guide evidence-aware text
    modulation, enabling adaptive vision-language fusion
    for fine-grained pest recognition.
    }
    \label{fig:pestweave_framework}
\end{figure*}

\subsection{Locality-Sensitive Hashing}
\label{sec:lsh}
Locality-Sensitive Hashing (LSH) maps feature vectors to discrete buckets such that nearby vectors have a higher collision probability than distant ones.
We employ Euclidean LSH (E2LSH)~\cite{datar2004locality} to initialize the grouping of visual tokens.
For $N$ projected visual tokens $v_i\in\mathbb{R}^{C}$, we first normalize each feature and compute its hash value as
\begin{equation}
  z_i=\frac{v_i}{\max(\lVert v_i\rVert_2,\epsilon)},\quad
  h_{\ell}(z_i)=
  \left\lfloor\frac{\xi_{\ell}^{\top}z_i+b_{\ell}}{r}\right\rfloor.
  \label{eq:e2lsh}
\end{equation}
Here, $\xi_{\ell}\sim\mathcal{N}(0,I_C)$ is a random projection vector, $b_{\ell}\sim\mathcal{U}(0,r)$ is a random offset, $r>0$ denotes the bucket width, and $\epsilon$ ensures numerical stability.
The floor operation quantizes the projected feature into an integer bucket index.

We apply $L_h$ independently sampled hash functions and combine their outputs into an integer sorting key:
\begin{equation}
  q_i^{(0)}=0,\qquad
  q_i^{(\ell)}=\beta_h q_i^{(\ell-1)}+h_{\ell}(z_i),
  \label{eq:lsh_key}
\end{equation}
where $\ell=1,\ldots,L_h$ and $\beta_h$ is a fixed integer base.
The random projections and offsets are sampled once and remain fixed during training.
For center initialization, we stably sort feature indices by $q_i^{(L_h)}$ and select $K_{\mathrm{cl}}$ approximately evenly spaced positions in the resulting order, where $1\leq K_{\mathrm{cl}}\leq N$ is the number of visual groups.
The corresponding normalized features serve as initial centers for subsequent spherical refinement.

\subsection{Network Framework}
\label{sec:framework}
As illustrated in Fig.~\ref{fig:pestweave_framework}, PestWeave integrates LSH-based semantic scanning, tri-token
augmentation, and evidence-guided vision--language
fusion. The input pest image is first processed by a
convolutional encoder and RWKV blocks, where
LSH-initialized grouping organizes visual tokens
for semantic scanning. Prototype, global, and
register tokens provide complementary context
for visual modeling. Meanwhile, the paired caption
is encoded by a frozen CLIP text encoder and
compressed into latent semantic tokens. Visual
prototypes guide the weighting and fusion of
these text tokens with visual features. Finally,
the fused representations are processed by a
Tail-Net and a subsequent classifier to produce
pest recognition predictions.

\subsection{Semantic-Enhanced Vision RWKV}
\label{sec:semantic_rwkv}
To capture both spatial relationships and visual similarity, we introduce a semantic scanning branch into Vision RWKV.
The module combines spatial and semantic interactions through a spatial mixer, followed by a channel mixer for feature transformation.
Given an input feature map $F\in\mathbb{R}^{H\times W\times C}$, we flatten it into a token sequence $X\in\mathbb{R}^{N\times C}$, where $N=HW$.

\noindent\textbf{Spatial RWKV Modeling.}
The input sequence is first normalized and locally mixed through a token-shift operation $\mathcal{S}_s$, then projected into Receptance, Key, and Value:
\begin{align}
  \widetilde X &= \mathcal{S}_s(\operatorname{LN}(X)),
  \quad R_s=\widetilde XW_R,
  \label{eq:spatial_shift} \\
  K_s &= \widetilde XW_K,\qquad V_s=\widetilde XW_V.
  \label{eq:spatial_projections}
\end{align}
Bidirectional WKV aggregates information across the sequence using position-dependent weights.
For a sequence of length $L$, its standard formulation is
\begin{align}
  \mathrm{wkv}_t
    &= \frac{\sum_{i\neq t}e^{-(|t-i|-1)w/L+k_i}v_i
                  +e^{u+k_t}v_t}
            {\sum_{i\neq t}e^{-(|t-i|-1)w/L+k_i}
                  +e^{u+k_t}},
  \label{eq:bidirectional_wkv} \\
  \mathrm{wkv} &= \mathcal{W}_L(K,V),
  \label{eq:wkv_operator}
\end{align}
where $k_i$ and $v_i$ are the $i$-th Key and Value tokens, $w$ and $u$ are learnable distance-weighting and self-token bias parameters, and all operations are applied channel-wise.
The spatial path performs successive horizontal and vertical scans:
\begin{align}
  V_h &= \mathcal{W}_N^{h}(K_s,V_s),
  \label{eq:horizontal_wkv} \\
  V_{\mathrm{sp}} &= \mathcal{T}^{-1}\!\left[
    \mathcal{W}_N^{v}(\mathcal{T}K_s,\mathcal{T}V_h)
  \right],
  \label{eq:vertical_wkv} \\
  Y_{\mathrm{sp}} &= \bigl(\sigma(R_s)\odot V_{\mathrm{sp}}\bigr)W_O,
  \label{eq:spatial_wkv}
\end{align}
where $\mathcal{T}$ converts row-major to column-major order and $\sigma$ denotes sigmoid gating.
The superscripts $h$ and $v$ denote horizontal and vertical scanning, respectively, with scan-specific WKV parameters.
These scans establish the spatial representation on which semantic modeling builds.

\noindent\textbf{LSH-Guided Semantic Grouping.}
Spatial proximity alone does not fully characterize the relationships among visually similar pest regions.
Let $z_i=\operatorname{Norm}(v_{s,i})$ denote the normalized $i$-th Value token.
The $K_{\mathrm{cl}}$ centers initialized by the LSH procedure in Section~\ref{sec:lsh} are refined through spherical clustering:
\begin{align}
  \ell_i^{(\nu)} &= \arg\max_{1\leq k\leq K_{\mathrm{cl}}}
    z_i^{\top}\mu_k^{(\nu)},
  \label{eq:semantic_assignment} \\
  \mu_k^{(\nu+1)} &= \operatorname{Norm}\!\left(
    \frac{1}{|\mathcal{C}_k^{(\nu)}|}
    \sum_{i\in\mathcal{C}_k^{(\nu)}}z_i
  \right),
  \label{eq:spherical_refinement}
\end{align}
where $\nu$ indexes refinement iterations, $\mathcal{C}_k^{(\nu)}=\{i:\ell_i^{(\nu)}=k\}$, and $\operatorname{Norm}$ denotes $\ell_2$ normalization.
The center update applies to nonempty groups; empty groups retain their previous centers.
Final assignments $\ell_i$ are obtained using the last updated centers, defining groups $\mathcal{C}_k=\{i:\ell_i=k\}$.
These groups define a permutation $\Pi$ that places their members contiguously, providing a content-adaptive order for semantic scanning.

\noindent\textbf{Tri-Token Semantic Scanning.}
To connect group-level appearance with image-wide context, we augment the semantic sequence with three types of auxiliary tokens: prototype, global, and register tokens.
For each nonempty group $\mathcal{C}_k$, its prototype summarizes the corresponding visual features, while the global token aggregates all spatial positions:
\begin{align}
  p_k &= \frac{1}{|\mathcal{C}_k|}
    \sum_{i\in\mathcal{C}_k}\widetilde x_i,
  \label{eq:visual_prototypes} \\
  g &= \frac{1}{N}\sum_{i=1}^{N}\widetilde x_i.
  \label{eq:global_token}
\end{align}
An empty group uses the global token as its prototype.
A learnable register token $r_{\mathrm{reg}}$ provides an additional context slot for joint aggregation.
Writing $P=[p_1,\ldots,p_{K_{\mathrm{cl}}}]^{\top}\in\mathbb{R}^{K_{\mathrm{cl}}\times C}$, the augmented sequence is
\begin{equation}
  Z=[\Pi\widetilde X;P;g^{\top};r_{\mathrm{reg}}^{\top}].
  \label{eq:augmented_sequence}
\end{equation}
The sequence is processed through the same projection and gating formulation as spatial RWKV:
\begin{equation}
  Y_e=\Bigl(\sigma(ZW_R)\odot
    \mathcal{W}_{M}^{e}(ZW_K,ZW_V)\Bigr)W_O,
  \label{eq:semantic_wkv}
\end{equation}
where $M=N+K_{\mathrm{cl}}+2$ is the augmented sequence length and the superscript $e$ denotes semantic scanning.
Within each block, the spatial and semantic paths share the projection weights $W_R$, $W_K$, $W_V$, and $W_O$, while using independent WKV parameters.
This joint scan enables interactions among visual tokens, group prototypes, and global context.

\noindent\textbf{Semantic Feature Integration.}
We first separate the semantic scanning output according to the augmented sequence:
\begin{equation}
  Y_e=[Y_x;Y_p;y_g^{\top};y_{\mathrm{reg}}^{\top}],
  \label{eq:semantic_outputs}
\end{equation}
where $Y_x$ contains the visual-token outputs in the reordered sequence, $Y_p$ contains the prototype outputs, and $y_g,y_{\mathrm{reg}}\in\mathbb{R}^{C}$ are the global and register outputs.
In this reordered sequence, the prototype outputs are broadcast to their group members to form $B_p$, while $y_g$ is broadcast to all visual positions to form $B_g$.
The semantic correction is then computed as:
\begin{align}
  \Delta_{\Pi}
    &= \lambda_x\odot Y_x+\lambda_p\odot B_p
     +\lambda_g\odot B_g,
  \label{eq:semantic_correction} \\
  Y &= Y_{\mathrm{sp}}+\Pi^{-1}\Delta_{\Pi}.
  \label{eq:semantic_residual}
\end{align}
Here, $\Pi^{-1}$ restores the original spatial arrangement after semantic integration.
For $q\in\{x,p,g\}$, the channel weights are parameterized as:
\begin{equation}
  \lambda_q=\tanh(\theta_q),\qquad \theta_q^{(0)}=0,
  \label{eq:semantic_scales}
\end{equation}
where $\theta_q\in\mathbb{R}^{C}$, yielding weights in $(-1,1)$ and an initially zero semantic correction.
The register participates in the joint scan, while its output is not directly propagated.
This residual formulation combines spatial structure with semantic relationships while allowing the semantic correction to enter gradually.

\noindent\textbf{Channel Mixing.}
The spatial--semantic representation is first integrated through a residual update:
\begin{equation}
  U=X+\gamma_s\odot Y.
  \label{eq:spatial_residual}
\end{equation}
The channel mixer then applies local mixing, nonlinear projection, and Receptance gating:
\begin{align}
  X_c &= \mathcal{S}_c(\operatorname{LN}(U)),
  \quad R_c=X_cW_R^c,
  \label{eq:channel_shift} \\
  K_c &= X_cW_K^c,\qquad V_c=\phi(K_c)W_V^c,
  \label{eq:channel_projections} \\
  Y_c &= \sigma(R_c)\odot V_c,\quad X'=U+\gamma_c\odot Y_c,
  \label{eq:channel_mix}
\end{align}
where $\phi(\cdot)=\operatorname{ReLU}(\cdot)^2$, $\mathcal{S}_c$ denotes token shifting, and $\gamma_s$ and $\gamma_c$ are learnable residual scales.
The resulting features combine spatial context, semantic interactions, and channel transformations for subsequent multimodal learning.

\begin{figure}[!t]
    \centering
    \includegraphics[width=\columnwidth]{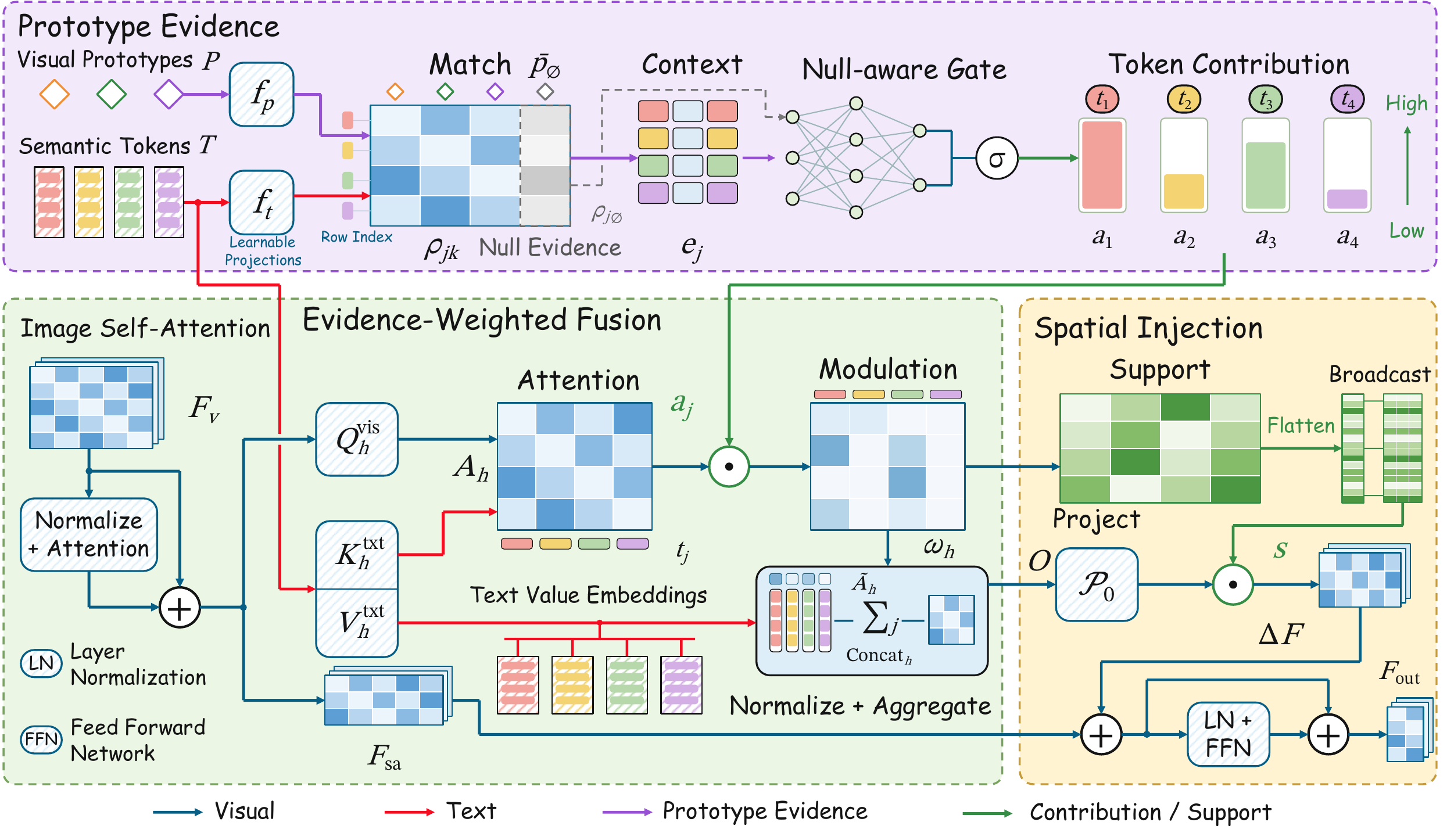}
    \caption{Overview of the proposed evidence-guided text modulation module. Visual prototypes estimate the support for textual semantics, guiding evidence-weighted cross-attention and spatially gated text injection for adaptive vision-language fusion.}
    \label{fig:text_modulation}
\end{figure}
\subsection{Evidence-Guided Vision-Language Fusion}
\label{sec:evidence_fusion}
Captions provide complementary semantic information
for pest recognition, while the visual support for
these semantics varies across images. As illustrated
in Fig.~\ref{fig:text_modulation}, we leverage visual
prototypes to assess the contributions of textual
semantics and regulate their interactions with
visual features. The resulting evidence guides both
semantic reweighting and spatially gated text
injection.

\noindent \textbf{Textual Semantic Encoding.}
Given the caption $c$ paired with an image, a frozen CLIP text encoder extracts its contextual representations.
A query-based compression module then summarizes the encoded sequence into a compact set of latent semantic tokens:
\begin{equation}
  E_c=E_{\mathrm{text}}(c),\qquad
  T=\mathcal{Q}(E_c;Q_0),
  \label{eq:text_semantic_encoding}
\end{equation}
where $Q_0$ contains learnable query tokens and $T=[t_1,\ldots,t_J]^{\top}\in\mathbb{R}^{J\times d_t}$.
We use $J=4$ semantic tokens to represent the paired caption.

\noindent\textbf{Prototype-Supported Evidence Estimation.}
The cluster-mean prototypes $P$ from the final RWKV block, constructed before semantic scanning as in Eq.~\ref{eq:visual_prototypes}, serve as visual support units for the textual semantics.
We map text tokens and visual prototypes into a common embedding space:
\begin{equation}
  \bar t_j=\operatorname{Norm}(f_t(t_j)),\quad
  \bar p_k=\operatorname{Norm}(f_p(p_k)),
  \label{eq:evidence_projections}
\end{equation}
Here, $f_t:\mathbb{R}^{d_t}\to\mathbb{R}^{d_e}$ and $f_p:\mathbb{R}^{C}\to\mathbb{R}^{d_e}$ are learned mappings into a common $d_e$-dimensional space, and $\operatorname{Norm}$ denotes feature-wise $\ell_2$ normalization.
A learnable normalized null key $\bar p_{\varnothing}\in\mathbb{R}^{d_e}$ provides an unmatched alternative when the visual prototypes offer limited support.
For each text token, support is distributed over the valid prototypes and the null key:
\begin{equation}
  \rho_{jk}=
  \frac{\exp(\tau_e\bar t_j^{\top}\bar p_k)}
       {\sum_{k^{\prime}\in\mathcal{K}\cup\{\varnothing\}}
        \exp(\tau_e\bar t_j^{\top}\bar p_{k^{\prime}})},
  \label{eq:prototype_support}
\end{equation}
where $k\in\mathcal{K}\cup\{\varnothing\}$, $\mathcal{K}=\{k:|\mathcal{C}_k|>0\}$ indexes nonempty groups of the final RWKV block, and $\tau_e>0$ is a learnable similarity scale.
The support assigned to real prototypes forms the visual context
\begin{equation}
  e_j=\sum_{k\in\mathcal{K}}\rho_{jk}\bar p_k.
  \label{eq:text_visual_context}
\end{equation}
We combine the textual representation, its visual context, their interaction, and the null support to predict a contribution weight:
\begin{equation}
  a_j=(1-\rho_{j\varnothing})\,
  \sigma\!\left(
    \operatorname{MLP}_{\vartheta}\bigl([
      \bar t_j;e_j;\bar t_j\odot e_j;\rho_{j\varnothing}
    ]\bigr)
  \right),
  \label{eq:text_contribution}
\end{equation}
where $\operatorname{MLP}_{\vartheta}:\mathbb{R}^{3d_e+1}\to\mathbb{R}$ is a learnable prediction function and $e_j\in\mathbb{R}^{d_e}$.
The weight $a_j\in[0,1]$ controls the contribution of the $j$-th semantic token according to its visual support.

\noindent\textbf{Evidence-Weighted Cross-Attention.}
For each fusion block, let $F_v\in\mathbb{R}^{N\times C}$ denote its input features; the first block receives the output of the visual RWKV network.
Image self-attention first updates their visual representation:
\begin{equation}
  F_{\mathrm{sa}}=F_v+\operatorname{SA}(\operatorname{LN}(F_v)).
  \label{eq:fusion_visual_update}
\end{equation}
The visual and textual features then generate queries, keys, and values for each attention head $h$:
\begin{equation}
  Q_h^{\mathrm{vis}} = \psi_q^h(F_{\mathrm{sa}}),  K_h^{\mathrm{txt}} = \psi_k^h(T),  V_h^{\mathrm{txt}} = \psi_v^h(T).
  \label{eq:fusion_qkv}
\end{equation}
Here, $Q_h^{\mathrm{vis}}\in\mathbb{R}^{N\times d_h}$ and $K_h^{\mathrm{txt}},V_h^{\mathrm{txt}}\in\mathbb{R}^{J\times d_h}$, where $d_h$ is the feature dimension of each head.
The projection functions include feature normalization and learned mappings, followed by normalization across visual positions for queries and across text tokens for keys and values.
The cross-attention distribution is
\begin{equation}
  A_h=\operatorname{softmax}\!\left(
    \eta_hQ_h^{\mathrm{vis}}
    \bigl(K_h^{\mathrm{txt}}\bigr)^{\top}
  \right),
  \label{eq:fusion_attention}
\end{equation}
Here, $A_h\in\mathbb{R}^{N\times J}$, $\eta_h$ is a learnable head-wise scaling parameter, and softmax operates over text tokens.
The contribution weights modulate this distribution while retaining its total support at each visual position:
\begin{equation}
  \omega_{hij}=A_{hij}a_j,\qquad
  m_{hi}=\sum_{j=1}^{J}\omega_{hij},
  \label{eq:weighted_text_support}
\end{equation}
where $i$ indexes visual positions.
The normalized weights and aggregated textual features are
\begin{align}
  \widetilde A_{hij}
    &= \frac{\omega_{hij}}{\max(m_{hi},\epsilon)},
  \label{eq:normalized_text_attention} \\
  O &= \operatorname{Concat}_{h=1}^{H_a}
    \bigl(\widetilde A_hV_h^{\mathrm{txt}}\bigr),
  \label{eq:aggregated_text_features}
\end{align}
Here, $H_a$ is the number of attention heads and $\epsilon$ ensures numerical stability.
Concat joins head outputs along the channel dimension, yielding $O\in\mathbb{R}^{N\times H_a d_h}$.
The normalized attention determines the relative semantic contributions, while $m_{hi}$ preserves their overall support for subsequent spatial modulation.

\begin{table*}[!t]
  \centering
  \caption{Quantitative comparison on three pest datasets. All metrics are percentages, with higher values indicating better performance.} 
\label{tab:main_results}
  \small
  \setlength{\tabcolsep}{3pt}

  \begin{tabular*}{\textwidth}{@{\extracolsep{\fill}}l*{12}{c}@{}}
    \toprule
    & \multicolumn{4}{c}{Li dataset} & \multicolumn{4}{c}{QianFSD} & \multicolumn{4}{c}{AgriInsect} \\
    \cmidrule(lr){2-5}\cmidrule(lr){6-9}\cmidrule(lr){10-13}
    Method & Acc. & Prec. & F1 & GM & Acc. & Prec. & F1 & GM & Acc. & Prec. & F1 & GM \\
    \midrule
    AlexNet~\cite{krizhevsky2012alexnet} & 60.80 & 59.50 & 59.42 & 60.57 & 54.87 & 55.95 & 54.87 & 46.64 & 64.14 & 65.16 & 64.33 & 62.31 \\
    VGG16~\cite{simonyan2015vgg} & 48.37 & 45.61 & 46.08 & 44.74 & 43.46 & 43.01 & 42.93 & 28.90 & 43.72 & 44.13 & 43.86 & 33.14 \\
    GoogLeNet~\cite{szegedy2015googlenet} & 54.89 & 53.87 & 54.26 & 54.44 & 48.76 & 50.21 & 49.15 & 46.91 & 52.19 & 53.72 & 52.74 & 52.92 \\
    ResNet-18~\cite{he2016resnet} & 77.98 & 78.73 & 77.91 & 77.33 & 69.03 & 70.73 & 69.44 & 68.50 & 80.29 & 81.95 & 80.90 & 82.00 \\
    ResNet-50~\cite{he2016resnet} & 71.48 & 71.74 & 71.15 & 69.58 & 68.09 & 69.74 & 68.68 & 67.12 & 80.34 & 81.68 & 80.74 & 82.91 \\
    ResNeXt-50~\cite{xie2017resnext} & 71.48 & 72.11 & 71.45 & 70.51 & 67.22 & 68.93 & 67.86 & 65.71 & 79.20 & 80.35 & 79.54 & 81.95 \\
    Swin-T~\cite{liu2021swin} & 75.23 & 74.67 & 74.73 & 75.73 & 68.02 & 70.71 & 69.05 & 68.64 & 82.24 & 83.28 & 82.48 & 84.93 \\
    UniFormer~\cite{li2022uniformer,li2023uniformer} & 71.56 & 70.77 & 70.72 & 71.63 & 76.50 & 78.10 & 76.93 & 76.56 & 65.75 & 67.53 & 65.98 & 59.01 \\
    SeaFormer~\cite{wan2023seaformer,wan2025seaformerpp} & 77.81 & 77.99 & 77.52 & 77.22 & 72.90 & 74.13 & 73.11 & 73.92 & 83.66 & 84.71 & 83.95 & 85.08 \\
    DFL~\cite{hu2024dfl} & 81.65 & 81.94 & 81.28 & 82.49 & 76.30 & 78.74 & 77.14 & 78.68 & 85.13 & 86.80 & 85.74 & 86.70 \\
    VMamba-B~\cite{liu2024vmamba} & 63.64 & 63.33 & 63.20 & 62.10 & 65.62 & 67.80 & 66.37 & 62.82 & 66.75 & 69.24 & 67.72 & 66.87 \\
    TransXNet~\cite{lou2025transxnet} & 76.90 & 76.31 & 76.27 & 76.59 & 79.44 & 80.36 & 79.56 & 81.86 & 80.67 & 81.81 & 81.04 & 81.95 \\
    LAViT~\cite{temiz2026lavit} & 57.21 & 57.36 & 56.85 & 52.21 & 58.81 & 60.07 & 58.93 & 51.38 & 53.10 & 55.07  & 53.63  & 44.32  \\
    HIVE-ViT~\cite{kim2026hive} & 63.05 & 64.24 & 63.26 & 59.38 & 60.48 & 62.23 & 60.83 & 56.27 & 64.90 & 66.64 & 65.70 & 63.63 \\
    PestVL-Net~\cite{li2026pestvl} & 88.49 & 89.36 & 88.52 & 88.56 & 86.72 & 87.53 & 86.86 & 87.14 & 90.15 & 90.88 & 90.19 & 90.51 \\
    \midrule
    \textbf{Ours} & \textbf{93.66} & \textbf{93.41} & \textbf{93.27} & \textbf{94.25} & \textbf{89.65} & \textbf{89.00} & \textbf{89.32} & \textbf{90.28} & \textbf{93.28} & \textbf{93.19} & \textbf{92.64} & \textbf{94.55} \\
    \bottomrule

  \end{tabular*}
\end{table*}

\providecommand{\pestAblationTableWidth}{0.9\columnwidth}
\begin{table}[!t]
    \centering
    \caption{Ablation study of PestWeave on QianFSD. All metrics are reported as percentages.}
    \label{tab:ablations}
    \small
    \setlength{\tabcolsep}{3pt}
    \renewcommand{\arraystretch}{1.12}
    \begin{tabular*}{\pestAblationTableWidth}{@{\extracolsep{\fill}}lcccc@{}}
        \toprule
        Variant & Acc. & Prec. & F1 & GM \\
        \midrule
        (I) w/o Text Fusion          & 81.11 & 82.51 & 81.49 & 83.88 \\
        (II) Unit Text Weights    & 83.78 & 85.28 & 84.11 & 84.19 \\
        (III) w/o Semantic Correction & 82.78 & 84.64 & 83.24 & 85.23 \\
        (IV) w/o Spherical Refinement        & 84.18 & 85.42 & 84.31 & 86.25 \\
        (V) w/o Tri-token Augmentation       & 84.85 & 86.36 & 85.18 & 87.62 \\
        (VI) w/o Null Option       & 85.11 & 86.28 & 85.33 & 87.75 \\
        \midrule
        \textbf{Full Model}       & \textbf{89.65} & \textbf{89.00} & \textbf{89.32} & \textbf{90.28} \\
        \bottomrule
    \end{tabular*}
\end{table}

\noindent\textbf{Spatially Gated Text Injection.}
We average the retained support across attention heads to obtain a spatial injection weight and use it to modulate the projected textual update:
\begin{equation}
  s_i=\frac{1}{H_a}\sum_{h=1}^{H_a}m_{hi},\qquad
  \Delta F=s\odot\mathcal{P}_0(O),
  \label{eq:spatial_text_injection}
\end{equation}
Here, $s=[s_1,\ldots,s_N]^{\top}\in\mathbb{R}^{N\times1}$ is broadcast across visual channels, and $\mathcal{P}_0$ maps each row of $O$ to $C$ visual channels with an initially zero output.
Thus, $\Delta F\in\mathbb{R}^{N\times C}$.
The fused representation is obtained through residual injection and feed-forward transformation:
\begin{equation}
  F_{\mathrm{mix}}=F_{\mathrm{sa}}+\Delta F,
  F_{\mathrm{out}}=F_{\mathrm{mix}}
    +\operatorname{FFN}(\operatorname{LN}(F_{\mathrm{mix}})).
  \label{eq:fusion_output}
\end{equation}
Preserving the support magnitude allows weak overall textual evidence to reduce the injected update even after attention normalization.
The same semantic tokens and contribution weights guide two successive fusion blocks, each computing its own attention and spatial injection weights.
In the following classification objective, $F_{\mathrm{out}}$ denotes the output of the second fusion block:
\begin{equation}
\hat{y} = \operatorname{softmax}\left(
H_{\mathrm{cls}}\bigl(\mathcal{T}_{\mathrm{tail}}(F_{\mathrm{out}})\bigr)
\right), \quad
\mathcal{L}_{\mathrm{cls}} = -\log \hat{y}_{y}.
\label{eq:classification}
\end{equation}
where $\mathcal{T}_{\mathrm{tail}}$ denotes the Tail-Net,
$H_{\mathrm{cls}}$ is the classification head,
and $y$ is the ground-truth class.

\section{Experiments}
\label{sec:experiments}
\setcounter{topnumber}{0}
\suppressfloats[t]
\AddToHookNext{shipout/after}{\setcounter{topnumber}{1}}
\begin{figure}[t]
    \centering
    \includegraphics[width=\columnwidth]{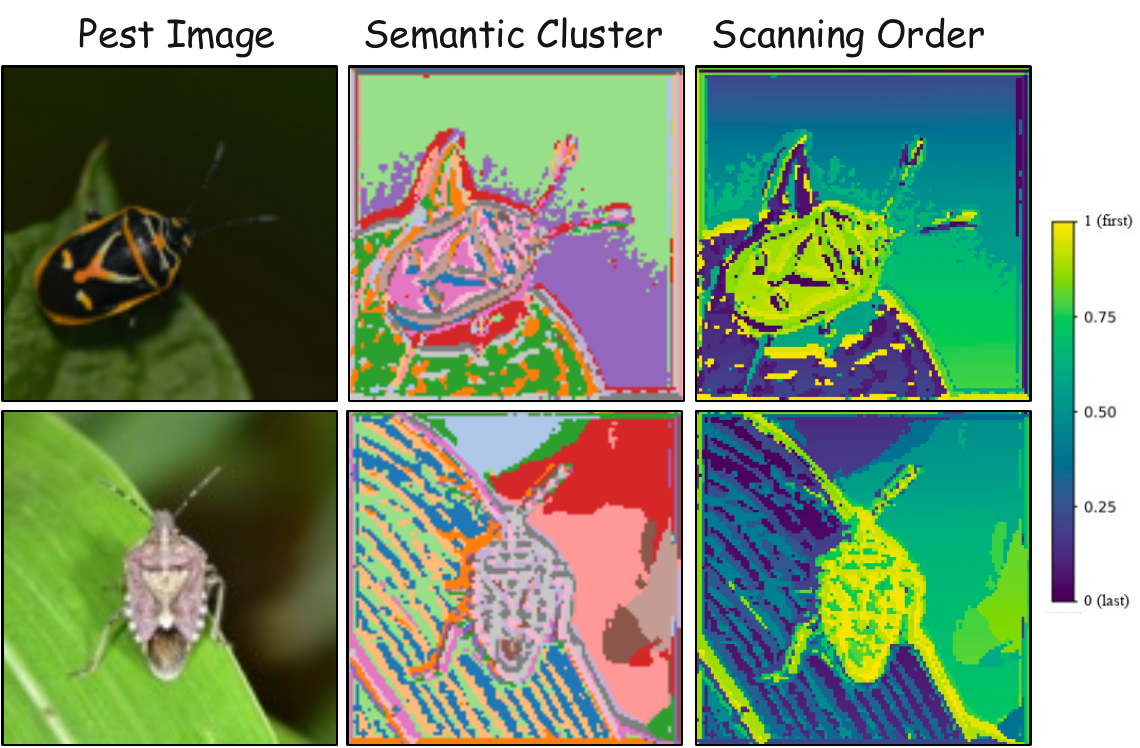}
    \caption{Visualization of semantic clustering and content-adaptive token scanning order.}
    \label{fig:semantic_clustering}
\end{figure}
\begin{figure}[t]
    \centering
    \includegraphics[width=\columnwidth]{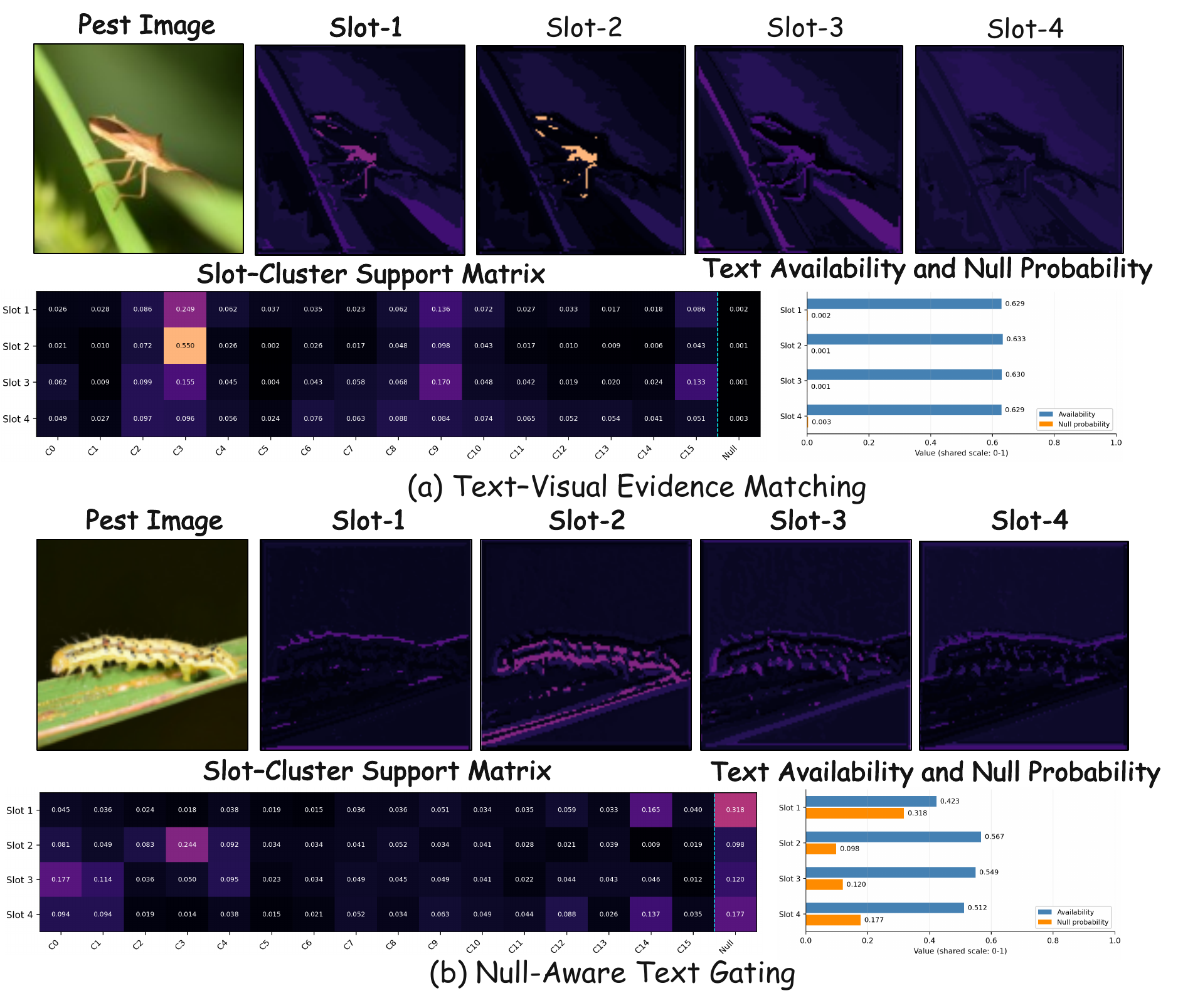}
    \caption{Visualization of prototype-supported textual evidence. (a) Learned semantic slots exhibit distinct support distributions over visual clusters. (b) Null-aware gating reduces the contribution of weakly supported semantic slots.}
    \label{fig:text_evidence}
\end{figure}

\subsection{Datasets}

We evaluate our method on three crop pest datasets with different category scales.
The Li dataset~\cite{li2020croppest} contains 5,629 natural-scene images from ten pest categories.
Its complex backgrounds pose challenges to visual feature extraction and pest recognition, enabling evaluation under background interference.
QianFSD~\cite{li2026pestvl} contains 7,054 images from 143 pest categories, while AgriInsect~\cite{li2026pestvl} comprises 9,452 images from 200 categories.
These datasets are partitioned into training, validation, and test sets with a 7:1:2 ratio. The latter two datasets encompass more diverse pest species and appearances, posing greater challenges for fine-grained recognition. For multimodal learning, we use publicly available image-level captions paired with individual samples. To avoid label leakage, captions are preprocessed to remove category names and other explicit label-identifying information. The corresponding processed captions are also provided at test time, ensuring a consistent multimodal evaluation protocol.


\subsection{Baseline Methods and Evaluation Metrics}

We compare PestWeave with 15 representative baselines on the Li dataset, QianFSD, and AgriInsect, as summarized in Table~\ref{tab:main_results}. The baselines include convolutional architectures such as AlexNet~\cite{krizhevsky2012alexnet}, VGG16~\cite{simonyan2015vgg}, GoogLeNet~\cite{szegedy2015googlenet}, ResNet-18/50~\cite{he2016resnet}, and ResNeXt-50~\cite{xie2017resnext}; attention-based and hybrid models including Swin-T~\cite{liu2021swin}, UniFormer~\cite{li2022uniformer,li2023uniformer}, SeaFormer~\cite{wan2023seaformer,wan2025seaformerpp}, and TransXNet~\cite{lou2025transxnet}; and the visual state-space model VMamba-B~\cite{liu2024vmamba}. We also consider pest-specific methods DFL~\cite{hu2024dfl}, LAViT~\cite{temiz2026lavit}, and HIVE-ViT~\cite{kim2026hive}, alongside the multimodal PestVL-Net~\cite{li2026pestvl}. Performance is assessed using accuracy (Acc.), precision (Prec.), F1-score (F1), and geometric mean (GM).


\subsection{Implementation Details}
We implement our model in PyTorch and train it on a single NVIDIA 4090 GPU. The model is optimized using stochastic gradient descent (SGD) with a momentum of 0.9, a batch size of 64, and an initial learning rate of 0.1.
We adopt the cross-entropy loss and train for 200 epochs on the Li dataset and 400 epochs on QianFSD and AgriInsect.
The trainable components are initialized from scratch, while the pre-trained CLIP~\cite{radford2021clip} text encoder remains frozen.

\providecommand{\pestAblationTableWidth}{0.9\columnwidth}

\subsection{Comparison with Baseline Methods}
Table~\ref{tab:main_results} presents the quantitative comparisons on three pest datasets. PestWeave consistently outperforms general-purpose architectures and pest-specific baselines across all evaluation metrics, demonstrating its effectiveness in fine-grained pest recognition. On the Li dataset, our method achieves 93.66\% accuracy and 94.25\% GM, exceeding PestVL-Net by 5.17 and 5.69 percentage points, respectively. Similarly, on QianFSD and AgriInsect, PestWeave improves accuracy by 2.93 and 3.13 percentage points, respectively, while achieving notable gains in F1-score and GM. These consistent improvements across datasets with varying category scales highlight the effectiveness of combining content-adaptive visual modeling with evidence-guided text modulation for discriminating pests with subtle morphological differences.

\subsection{Ablation Studies}
We conduct ablation studies on QianFSD to evaluate the contributions of individual components, as summarized in Table~\ref{tab:ablations}. All variants retain the image self-attention and feed-forward layers in the fusion blocks to ensure consistent visual modeling capacity.

\noindent\textbf{Textual Information and Evidence Weighting.}
Variant (I) disables text injection while retaining the visual processing operations, reducing accuracy from 89.65\% to 81.11\%. Variant (II) fixes all semantic contribution weights to one while preserving text fusion, resulting in 83.78\% accuracy. These results demonstrate the benefits of incorporating textual information and regulating its contributions according to visual evidence.

\noindent\textbf{Semantic Visual Organization.}
Variant (III) removes semantic WKV correction while retaining spatial scanning and visual prototypes for text modulation, leading to a 6.87-point accuracy drop. Variant (IV) retains LSH initialization and cosine assignment but removes iterative spherical refinement, reducing accuracy to 84.18\%. Variant (V) excludes prototype, global, and register tokens from semantic scanning while preserving cluster prototypes for text evidence estimation, achieving 84.85\% accuracy. These results highlight the contributions of semantic scanning, clustering refinement, and tri-token augmentation to discriminative visual representation learning.

\noindent\textbf{Unmatched Visual Evidence.}
Variant (VI) removes the null-evidence option while retaining real-prototype matching and the learned contribution gate. Accuracy decreases to 85.11\%, with corresponding declines in F1-score and GM. This demonstrates the benefit of explicitly modeling unmatched visual evidence to reduce the influence of unsupported textual semantics.

\subsection{Qualitative Analysis}
Fig.~\ref{fig:semantic_clustering} visualizes semantic cluster assignments and the corresponding scanning orders on representative pest images. The results illustrate how content-adaptive grouping organizes visually related regions into semantic clusters and brings their tokens closer in the scanning sequence. Fig.~\ref{fig:text_evidence} further visualizes prototype-supported textual evidence through slot--cluster support matrices, spatial evidence maps, and null-aware availability weights. Different semantic slots exhibit distinct support distributions, with lower null probabilities indicating stronger visual support. Conversely, higher null probabilities correspond to reduced availability, demonstrating adaptive regulation of textual contributions. The spatial maps reflect cluster-level evidence rather than word-level localization or actual text injection strength.

\section{Conclusion}
In this work, we presented PestWeave, a multimodal framework for fine-grained pest recognition that connects semantic scanning with evidence-guided text modulation through visual prototypes. By introducing LSH-assisted token grouping and semantic scanning, PestWeave strengthens interactions among visually related regions while preserving spatial structure. We further incorporate cluster prototype, global, and register tokens to enhance contextual modeling and refine visual representations. For cross-modal fusion, we employ prototype-supported evidence estimation with a null-evidence option to assess the applicability of textual semantics, jointly regulating their relative contributions and overall injection strength to reduce interference from unsupported descriptions. Extensive experiments on multiple datasets demonstrate the effectiveness of PestWeave in fine-grained pest learning, highlighting the complementary benefits of content-adaptive visual organization and evidence-guided text modulation.

{
    \small
    \bibliographystyle{ieeenat_fullname}
    \bibliography{main}
}
\end{document}